\documentclass[runningheads]{llncs}
\usepackage{graphicx}
\usepackage{cite}
\usepackage{amsmath}
\usepackage{bm}    
\usepackage{varioref}     
\usepackage{float} 

\usepackage{array}
\newcolumntype{P}[1]{>{\centering\arraybackslash}p{#1}}

\usepackage[shortlabels]{enumitem}

\usepackage{xcolor}

\usepackage{multirow}

\usepackage{arydshln}

\usepackage{arydshln} 

\begin{document}

%
\title{Creating Unbiased Public Benchmark Datasets with Data Leakage Prevention \\for Predictive Process Monitoring}
\titlerunning {Benchmark Datasets for Predictive Process Monitoring}
%
%
\author{Hans Weytjens\and
Jochen De Weerdt}
\authorrunning{H. Weytjens, J. De Weerdt}
%
\institute{Research Centre for Information Systems Engineering (LIRIS),\\ KU Leuven, Leuven, Belgium\\
\email{\{hans.weytjens,jochen.deweerdt\}@kuleuven.be}}
\maketitle              
\begin{abstract}
Advances in AI, and especially machine learning, are increasingly drawing research interest and efforts towards predictive process monitoring, the subfield of process mining (PM) that concerns predicting next events, process outcomes and remaining execution times. Unfortunately, researchers use a variety of datasets and ways to split them into training and test sets. The documentation of these preprocessing steps is not always complete. Consequently, research results are hard or even impossible to reproduce and to compare between papers. At times, the use of non-public domain knowledge further hampers the fair competition of ideas. Often the training and test sets are not completely separated, a data leakage problem particular to predictive process monitoring. Moreover, test sets usually suffer from bias in terms of both the mix of case durations and the number of running cases. These obstacles pose a challenge to the  field's progress. The contribution of this paper is to identify and demonstrate the importance of these obstacles and to propose preprocessing steps to arrive at unbiased benchmark datasets in a principled way, thus creating representative test sets without data leakage with the aim of levelling the playing field, promoting open science and contributing to more rapid progress in predictive process monitoring.

\end{abstract}
%

%
\section{Introduction}\label{sec:intro}
Process mining analyzes event data logged by information systems with the goal of process discovery, process conformance checking and process enhancement. Predictive process monitoring is an important sub-field of process mining and concerns predicting next events, process outcomes and remaining execution times. The field is enjoying increased interest and is progressing thanks to opportunities offered by developments in AI and machine learning and the availability of data. Consequently, an increasing number of papers concerning predictive process monitoring are being published. Studying them, however, we identified three major obstacles to further progress in the field. 

The first obstacle involves the use of different datasets that complicates or even impedes the comparison of results between papers. Even though many researchers use publicly available datasets, differences in their pruning and training/test set splits lead to significantly diverging results. Somewhat unfairly, some of these preprocessing decisions may have been guided by domain knowledge unavailable to all researchers. The second obstacle concerns the proper splitting of training and test sets. When using standard temporal splitting or cross-validation techniques, a number of cases will have prefixes in both the training and test set which is a form of data leakage and affects prediction performance. Most papers do not remedy this issue. For correct science, however, predictive process monitoring problems require a specific data split technique. The third obstacle relates to two forms of bias that are often ignored: when crudely determining start and end times for training and test sets, the number and average length of running cases at any given time may be greatly impacted at the chronological beginning and ending of these training and test sets.

To overcome these three obstacles, we believe predictive process monitoring needs benchmark datasets with predefined targets, training and test sets. With such datasets available, researchers could process the training sets at will while having to test their results on the given test sets. Benchmark datasets have always played an important role in major machine learning areas such as natural language\footnote{\url{https://gluebenchmark.com/}}, reinforcement learning\footnote{\url{https://arxiv.org/abs/2004.07219}}, image recognition, etc. They allow the research community to compare different proposed methods and spur the development of new state-of-the-art ideas, often in the form of competitions. Well-known examples in the field of image recognition include  --sorted by increasing size and complexity-- MNIST\footnote{\url{http://yann.lecun.com/exdb/mnist/}} (60,000 black-white handwritten digit images), CIFAR-10\footnote{\url{https://www.cs.toronto.edu/~kriz/cifar.html}} (60,000 color images in 10 classes) and ImageNet\footnote{\url{https://image-net.org/}} (14 million images in 20,000 classes), progressively used as the field matured.

The contribution of this paper is twofold. First, we want to expose the three obstacles mentioned above by investigating nine of the most commonly used public process mining datasets. Second, based on this analysis, we derive a set of proposals on how to design benchmark datasets for predictive process monitoring. These will ensure fairness, enable reproducibility and facilitate progress in the field. The code to implement our proposals is publicly available\footnote{\url{https://github.com/hansweytjens/remaining-time-benchmarks}}.

The next section refers to influential process outcome prediction research and relates it to the three obstacles. We then detail our approach in Section~\ref{sec:methodology} and describe our experimental setup in Section~\ref{sec:setup}. We derive preprocessing steps that convert public datasets into unbiased benchmark datasets in Section~\ref{sec:measures}. Using a simple convolutional neural network (CNN), Section~\ref{sec:results} exposes the impact of dataset design decisions before concluding in Section~\ref{sec:conclusions}.

\section{Related work}
When exploring some of the most influential original papers and surveys on outcome \cite{teinemaa, kratsch}, next event \cite{tax, evermann, camargo} or remaining time \cite{tax, survey, dataaware} prediction, the first obstacle of poor comparability becomes immediately apparent. Even where researchers work with the same datasets, direct comparisons of their results prove impossible. For example, the classes for the BPIC\_2011 and Traffic datasets are defined differently by Teinemaa et al.\cite{teinemaa} and Kratsch et al.\cite{kratsch}. In next event prediction, Tax et al. \cite{tax} consider only events of type ``complete'' whilst Evermann et al. \cite{evermann} create new features by concatenating the ``activity'' and ``resource'' features. In the remaining time prediction problem, Tax et al.\cite{tax} and Verenich et al. \cite{survey} use different ratios ($\frac{2}{3}$/$\frac{1}{3}$ and 80\%/20\% respectively) when chronologically splitting training and test sets, whereas Polato et al. \cite{dataaware} use fivefold cross-validation. In their literature overview study, Neu et al. \cite{neu} acknowledge the latter problem and propose to ``use a unified evaluation approach for further publications.'' Other than Verenich et al. \cite{survey} recognising (but also accepting) the data leakage problem when naively splitting training and test sets, we found no mention of the second obstacle. To the best of our knowledge, no research diagnoses the third obstacle of bias.

\section{Approach}\label{sec:methodology}
To counter the first obstacle's lack of comparability, we will formulate a standardised preprocessing approach to construct benchmark datasets in Section~\ref{sec:measures}. Along the steps in this preprocessing procedure, we will observe and remedy both the second obstacle of data leakage between test and training sets and the third obstacle of bias as they appear. We will use nine datasets to analyze the problems and illustrate the preprocessing steps and their effects in the context of remaining time prediction. We will point out where classification or next-event prediction deviate from remaining time to ensure our approach applies to all predictive process monitoring problems.
Before doing so, we first outline the four main guiding principles that steer this work:
\begin{itemize}
\item \textbf{A principled approach}: increases transparency and enables reproducibility to other datasets / problems,
\item \textbf{Proximity to real-life}: creates datasets that resemble the original underlying business problems as much as possible,
\item \textbf{Enabling of good science}: (only) where required, we deviate from the first two principles for the benefit of science,
\item \textbf{Domain agnosticism}: as domain knowledge is often sparse and unevenly distributed amongst researchers, we refrain from using it altogether.
\end{itemize}

We demonstrate how different preprocessing impacts prediction results. It is rather unconventional to illustrate the importance of the problem towards the end of a paper; however, in our case, it offers the benefit of having clearly-defined preprocessing steps that each generate different variants of the original datasets on which we ran simple convolutional network to predict remaining times. 

\section{Experimental setup}\label{sec:setup}
\vspace*{-.1cm}
\subsection{Preliminaries}
In predictive process monitoring, datasets are event logs describing processes, often called \textit{cases}. These cases consist of \textit{events}. A number of \textit{features} describe these cases and events. In remaining time prediction problems, every event is associated with a \textit{target} feature containing the remaining time until completion of the case.  A \textit{prefix} is an ongoing case, with the \textit{prefix length} its number of completed events. The learning problem is to train an algorithm on a training dataset containing events, described by their features and organised in prefixes labelled with targets, with the goal of predicting the targets of unseen prefixes. \\
We distinguish two phases in the preprocessing process. We will restrict the use of the term \textit{preprocessing} to refer to the first phase steps that are only performed once and lead to the construction of unique benchmark training and test sets. The second phase involves further steps, such as feature selection and standardisation taken by individual researchers to prepare these fixed training and test sets to train and test their models. This latter phase is not subject of this paper and belongs to the area where researchers and ideas compete.

\vspace*{-.1cm}
\subsection{Datasets}
 We used five publicly available datasets from the 4TU.ResearchData repository. BPIC\_2012\footnote{https://data.4tu.nl/articles/dataset/BPI\_Challenge\_2012/12689204} contains logs of a loan application process at a Dutch bank. BPIC\_2015 \footnote{https://data.4tu.nl/collections/BPI\_Challenge\_2015/5065424} is a collection of building permit applications in five Dutch municipalities, which we concatenated into one dataset. BPIC\_2017\footnote{https://data.4tu.nl/articles/dataset/BPI\_Challenge\_2017/12696884} is a richer, cleaner and larger set of logs of a Dutch bank loan application process. BPIC\_2019\footnote{https://data.4tu.nl/articles/dataset/BPI\_Challenge\_2019/12715853}, while comparable in size, has much shorter cases and concerns a purchase order handling process. BPIC\_2020\footnote{https://data.4tu.nl/collections/BPI\_Challenge\_2020/5065541} is a collection of five smaller datasets related to travel administration at a university. The five subsets are records of processes covering domestic declaration documents (Domestic Declarations), international declaration documents (Intl. Declarations), pre-paid travel costs and requests for payment (Payments), travel permits (Permits) and expense claims (Travel Costs). Our target for all of these datasets was defined as the fractional number of days until case completion. Table~\ref{tab:data} provides some key statistics of these datasets.

\vspace*{-1mm}
\begin{table}
\centering
\vspace*{-1mm}
\caption{Statistics of the used datasets$^*$.}\label{tab:data}
\scalebox{0.999}{
\begin{tabular}{|l||c|c|c|c|c|c|c|c|}
\hline
Dataset & Nr.      & Nr.      & Median  & Avg.     &Median   & Mean & Max & Dataset \\
              & cases  &events& nr.           & nr.        & nr.           & nr.       & nr.    &  nr.\\
              &            &            & events    & events & days       & days   &days  &  days\\
\hline
BPIC\_2012  &12,183&228,873&9&18.8&0.5&7.8&137.2&152\\
BPIC\_2015  &5,641&262,328 &45&46.5&68.4&100.8&1,512&1,617 \\
BPIC\_2017 &31,497&1,201,390 &35&38.1&19.1&21.9&286.1& 397\\
BPIC\_2019 &251,465&1,587,810&5&6.3&64.0&69.5&380.0& 830\\
\hdashline
BPIC\_2020:&& &&&&&& \\ 
Domestic. Declarations&6,087&56,359&5&9.3&14.3&101.7&735.2& 773\\
Intl. Declarations&1,440&71,735&13&49.8&416.1&382.9&781.7&1,010 \\
Payments&6,593&35,046&5&5.3&7.9&11.1&1,238&711 \\
Permits&7,063&86,560&11&12.3&71.7&87.1&502.4&1121 \\
Travel Costs&2,097&18,238&8&8.7&24.0&36.7&325.0& 749\\

\hline
\multicolumn{9}{l}{\small *Outliers already removed as per Subsection~\ref{subsec:chronoutliers}.} \\
\end{tabular}
}
\vspace*{-1mm}
\end{table}
\vspace*{-1mm}

\vspace*{-.1cm}
\subsection{Models}
To illustrate the relevance of the preprocessing choices to construct datasets, we calculated the mean absolute error (MAE) of a CNN model on various preprocessing versions of the datasets. For each dataset, we included one categorical (``activity'') and one range (``elapsed'': time since case start) feature. We used embedding to feed the categorical feature in the models and reduced its dimensionality to the square root. Two convolutional layers followed the embedding layer, each with 40 kernels of size three and a max-pooling operation with a window of size three. The stride size for both the kernels and pooling was one. Two dense layers of sizes 100 and ten and with a ReLu activation completed the architecture. We reduced overfitting with early stopping using a validation set of the 20\% latest-starting cases in the training sets and a patience of 30. The sequence length was ten, the batch size 2048.

\section{Constructing benchmark datasets}\label{sec:measures}

\vspace*{-.1cm}
\subsection{Target definition}\label{subsec:target}
Our targets are defined as the remaining time between an event's time stamp and the case's last time stamp. For classification tasks, determining targets requires weakening the domain agnosticism principle and recourse to an authority for a final decision. To cope with unbalanced class distributions, we propose using area under the receiver operating characteristics curve (AUC-ROC) as a metric, rather than the potentially misleading accuracy.

\vspace*{-.1cm}
\subsection{Elimination of chronological outliers and duplicates}\label{subsec:chronoutliers}

\begin{figure}[h]
\vspace*{-1mm}
\center
\includegraphics[width=1\textwidth]{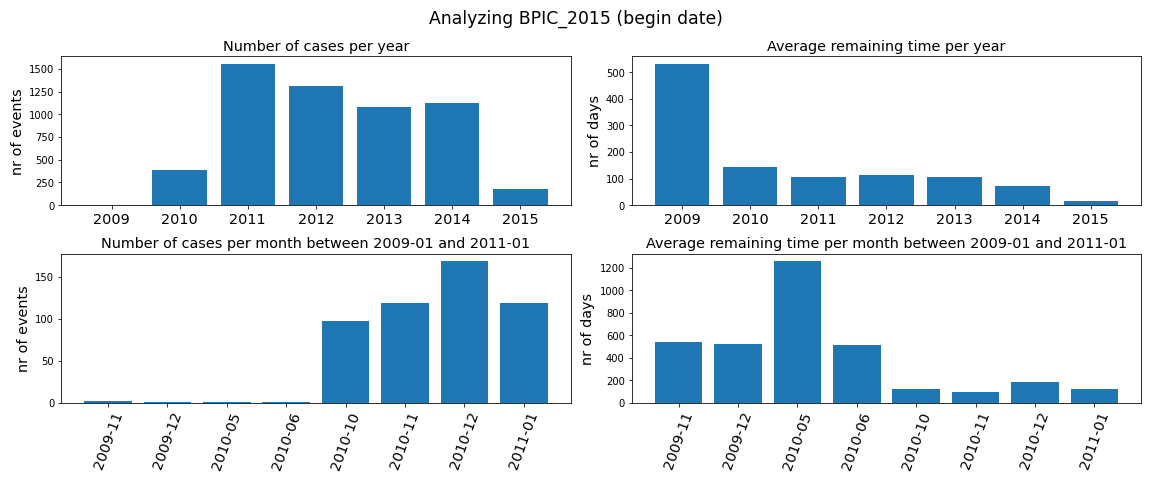}
\caption{BPIC\_2015 contains remote, faulty outliers at the beginning of the dataset.} \label{fig:analysis_2015}
\vspace*{-1mm}
\end{figure}

The BPIC\_2015 dataset contains cases starting in the period 2009-2015. As is shown in the top row on Fig.~\ref{fig:analysis_2015}, the cases in the year 2009 appear aberrant as they are few in number and last much longer than cases in the following years. When zooming in on cases starting before 2011-01-31st (bottom row), we see that the logs start behaving ``normally'' as of October 2010. In the following, we only included cases starting after Oct 1st, 2010. We found similar outliers in BPIC\_2019.
\vspace*{-1mm}
\vspace*{-1mm}

Similar patterns arose for the case's ending times at the end of all datasets involved. To remove those outliers, we imposed new end dates. An overview can be found in Table~\ref{tab:outliers}.

\vspace*{-1mm}
\begin{table}[H]
\centering
\vspace*{-1mm}
\caption{Start and end dates before and after eliminating outliers.}\label{tab:outliers}
\scalebox{0.999}{
\begin{tabular}{|l||c|c|c|c|}
\hline
\multirow{2}*{Dataset} &\multicolumn{2}{c|}{Original dataset}&\multicolumn{2}{c|}{After removing outliers}\\
\cline{2-5}
&First event & Last event & Start in/after      & End before/in      \\
            
\hline
BPIC\_2012  &2011-10&2012-03&-&2012-02\\
BPIC\_2015  &2009-11 &2015-07&2010-10&- \\
BPIC\_2017 & 2016-01&2017-02&-&2017-01\\
BPIC\_2019 & 1948-01&2020-04&2018-01&2019-02\\
\hdashline
BPIC\_2020:& &&-& \\ 
Domestic Declarations& 2017-01&2019-06&-&2019-02 \\
Intl. Declarations&2016-10 &2020-05&-&2019-07\\
Payments&2017-01 &2019-08&-&2018-12 \\
Permits&2016-10 &2021-08&-& 2019-10\\
Travel Costs&2017-01 &2019-02&-&2019-01\\

\hline
\end{tabular}
}
\vspace*{-1mm}
\end{table}
\vspace*{-1mm}

A dataset may also contain faulty outliers. These are not detectable without domain knowledge and therefore we did not attempt to remove them. The elimination of duplicates requires no further discussion. None were found in our datasets. For clarity, all tables and graphs (outside of this subsection) in this paper, including the preceding Table~\ref{tab:data}, rely on datasets without chronological outliers.

\vspace*{-.1cm}
\subsection{Debiasing the end of the dataset}\label{subsection:phaseout}
\begin{figure}[h]
\vspace*{-1mm}
\center
\includegraphics[width=1\textwidth]{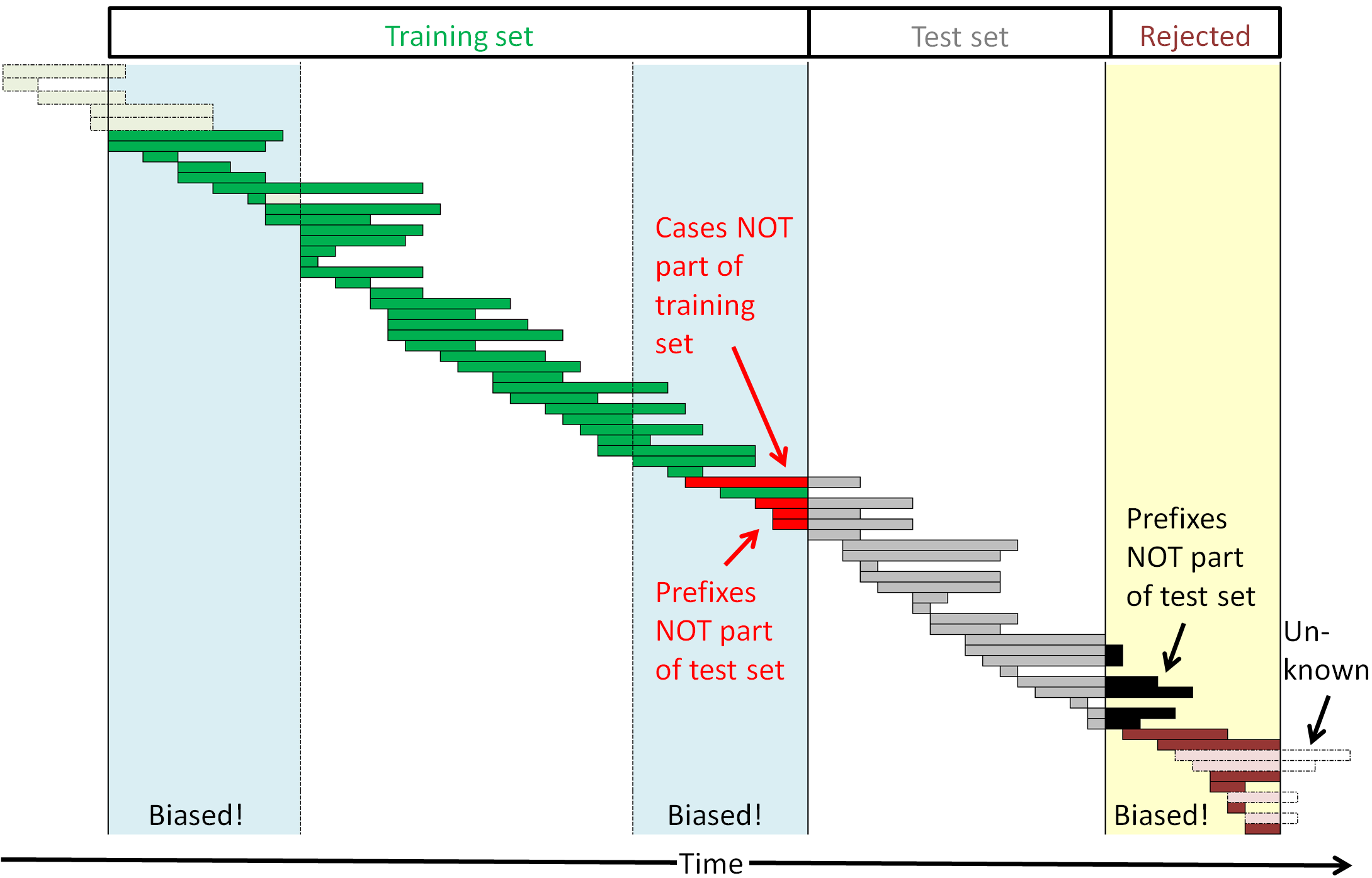}
\caption{An artificial dataset with cases ranked by starting time.} \label{fig:traintest}
\vspace*{-1mm}
\end{figure}

In order to know their targets, a dataset should only contain completed cases. This introduces two forms of bias towards the end of the dataset (third obstacle). First, cases' average durations will inevitably decrease as their start dates approach the dataset's end. Fig.~\ref{fig:traintest} visualises this with an artificial example. The light maroon cases are dropped as they are incomplete. This biases the yellow zone comprising the remaining cases towards shorter ones. Moreover, the number of running cases no longer reflects the underlying reality, which is the second form of bias. It prohibits the use of inter-case variables to compute for example the load of a resource dealing with the cases. The width of this yellow zone equals the duration of the dataset's longest case. To debias the ending of the dataset, we reject all maroon cases as well as the (black) \textit{prefixes} ending in the yellow zone. Cases for which we removed the larger (black) \textit{prefixes}, still remain in the dataset with their shorter (grey) \textit{prefixes} whose targets are known.

\vspace*{-1mm}
\vspace*{-1mm}

This step also automatically removes potentially incomplete cases at the moment of the last timestamp in the original dataset. Note that one (possibly faulty) long case in the dataset could suffice to severely reduce the training set size or even annihilate it when debiasing the dataset's end. This problem will be tackled below in Subsection~\ref{subsec:longcases}. Strictly spoken, next event prediction problems don't require completed cases. Leaving the light maroon cases' known prefixes in the dataset would nevertheless cause the first bias towards shorter cases.

\vspace*{-.1cm}
\subsection{Choice of test set}
Some authors (e.g.\cite{dataaware, evermann}) use cross-validation for predictions (not just whilst training) to make their models robust against concept drift. However, some of that concept drift can be attributed to the end-of-dataset bias we described and remedied above. Moreover, concept drift happens in real-life situations too. It may be even learnable to some extent. As the strict separation of training and test sets (second obstacle) is also more difficult to implement in a cross-validation setting, we recommend using a classic temporal training/test split for all of these reasons. Larger test sets are more likely to be representative for the whole dataset, but will reduce the size of the corresponding training set and carry a higher risk of real concept drift. We propose that test sets consist of the last 20\% of the datasets' cases chronologically. 

\vspace*{-.1cm}
\subsection{Temporal splitting}\label{subsection:strictts}
To isolate a test set, many authors rank cases by the timestamp of their first event and determine a separation time, with the x\% last cases starting after that time going to the test set and the remainder constituting the training set. From a real-life perspective, the logical flaw is that the outcome of incomplete cases in the training set (red/grey in Fig.~\ref{fig:traintest}) is not known at that separation time. Moreover, some cases from the training set and test will be running concurrently and influencing each other. This is the second obstacle mentioned in the introduction. To counter it, we recommend to always apply ``strict temporal splitting'' by only retaining in the training set those cases that are completed before the separation time. These are colored in green in Fig.~\ref{fig:traintest}. As a result, the training set will contain two biased zones (light blue), at its beginning and end. It is left to researchers to debias these zones as they see fit. A compelling reason to make this imperative is not present.

\begin{figure}[h]
\vspace*{-3mm}
\center
\includegraphics[width=1\textwidth]{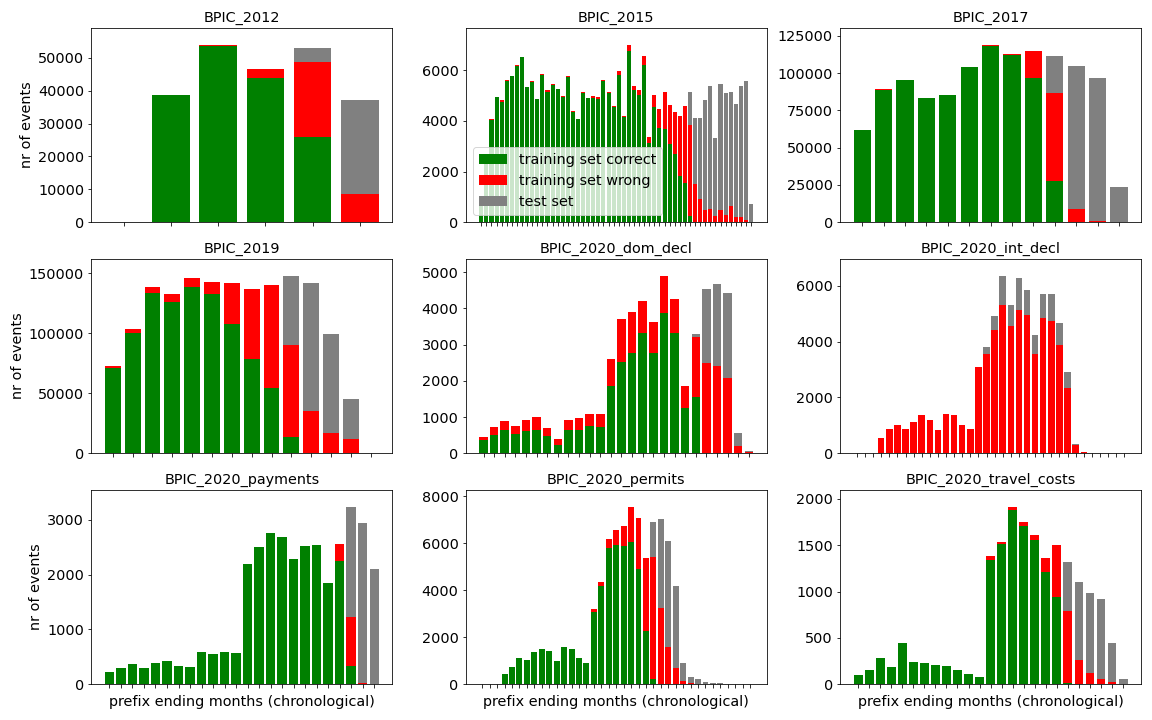}
\caption{Grey bars represent test set events (20\%). Green bars are events in training set cases ending before the separation time (strict temporal splitting), red ones belong to cases ending after that separation time (regular temporal splitting). As debiasing was not done to retain sufficient samples, the bias at the datasets' ends is clearly visible.} \label{fig:events_splits}
\vspace*{-.5cm}
\end{figure}

Depending on the duration of the cases in relation to the time span covered by the dataset, the difference between strict and regular temporal splitting can be rather significant, as demonstrated in Fig.~\ref{fig:events_splits}.

\vspace*{-.1cm}
\subsection{Debiasing the beginning of the test set}\label{subsection:balanced}

We now face both obstacle three biases at the beginning of the test set. Of the (red-grey) cases that start before the separation time but end after it, we also include in the test set those (grey) \textit{prefixes} that end after the separation time. This restores balance both in terms of the case lengths and the number of running cases. It also better resembles real-life settings where a maximum of available data at any given time would be considered.

\vspace*{-.1cm}
\subsection{Removal of long cases}\label{subsec:longcases}
Applying our third guideline, we suggest discarding extremely long cases from the dataset in the interest of the dataset's survival when removing biases at the end of the dataset as described in Subsection~\ref{subsection:phaseout}. This clearly changes the nature of the original dataset, but we believe that leaving the biases in would have a more negative impact on  the applicability of the models being developed in that scenario. For every dataset, Fig.~\ref{fig:sim_case_lengths_short_testset} plots the size of the training and test sets (in number of events) for different maximum case durations for two test-case-share scenarios (blue: 10\% and orange: 20\% of the total remaining dataset). debiasing at the end of the dataset and beginning of the test set and strict temporal splitting were performed. Not excluding the longest cases from them would drastically, often fatally shrink the datasets as can be seen on the right side of the graphs. Some degree of pragmatism is called for: we propose removing up to, but no more than, 5\% of the longest-lasting cases to find the case duration within the window that corresponds with the largest training set (in number of cases) as shown by the red lines in Fig.~\ref{fig:sim_case_lengths_short_testset}. We, thus, obtain the training and test sets in Tables~\ref{tab:proposed1} and ~\ref{tab:proposed2}. The (quasi) disappearance of the training sets for BPIC\_2020 Domestic and Intl. Declarations disqualifies them for serious research.

\section{Model predictions}\label{sec:results}
\begin{figure}[h]
\vspace*{-.1cm}
\center
\includegraphics[width=1\textwidth]{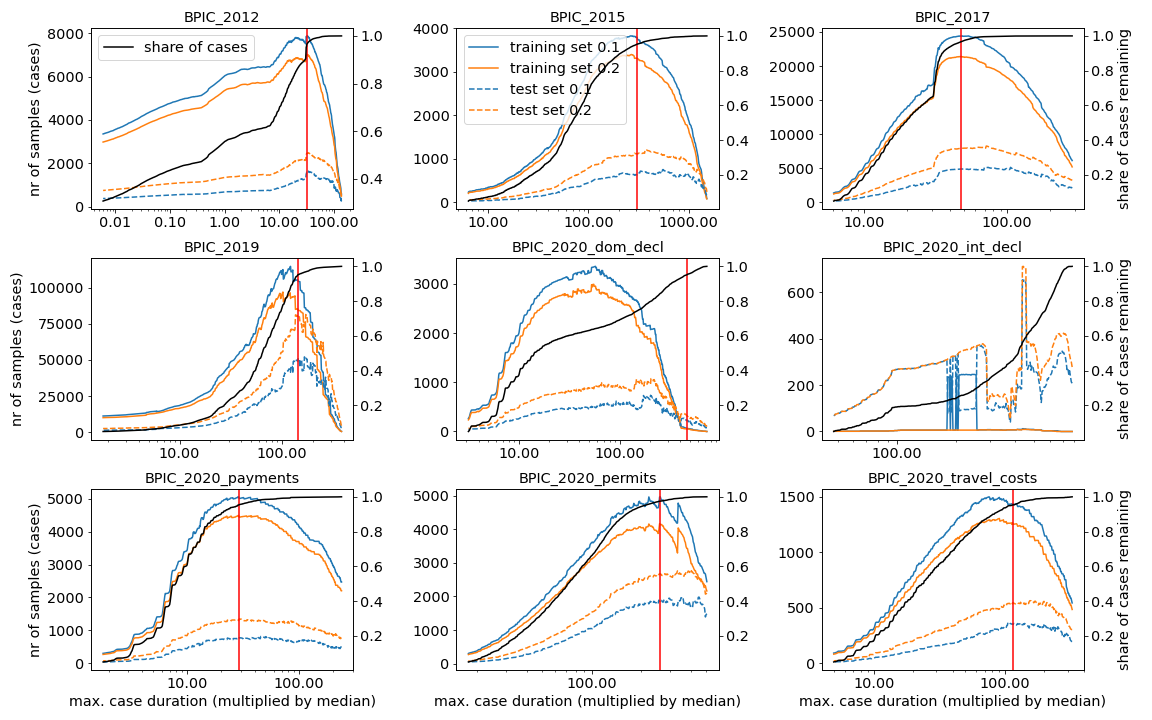}
\caption{Size of training and test sets (10\%/20\%) in function of maximal case lengths. Debiasing and strict temporal splitting applied. Vertical red lines show our proposals.} \label{fig:sim_case_lengths_short_testset}
\vspace*{-.1cm}
\end{figure}

\begin{figure}[H]
\vspace*{-.1cm}
\center
\includegraphics[width=1\textwidth]{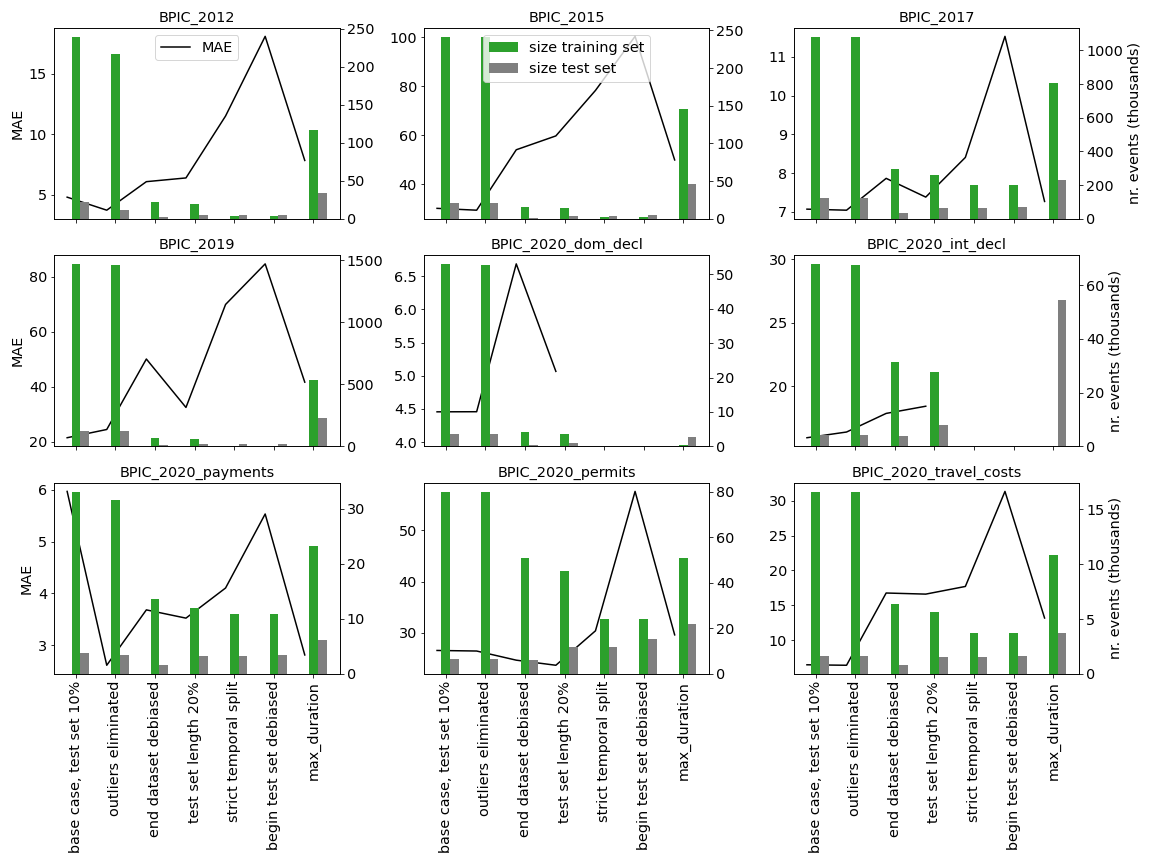}
\caption{Prediction results (MAE) vary significantly in function of the chosen training and test data sets.} \label{fig:mae}
\vspace*{-.1cm}
\end{figure}

Fig.~\ref{fig:mae} visualises the impact of the preprocessing decisions on predictions. For every dataset, we ran seven simulations. The first simulation concerned a base case with a train/test split of the first 90\% and last 10\% starting cases. For each of the following six simulations, we successively introduced the measures discussed in Section~\ref{sec:measures}. The maximal durations used in the last simulation correspond to the red vertical lines in Fig.~\ref{fig:sim_case_lengths_short_testset}. The resulting quality of the predictions from the CNN model, expressed as MAE, is represented by the black lines (left vertical axis). The obtained sizes of training (green bars) and test set (grey bars) are calibrated to the right vertical axis. 

\vspace*{-5mm}
\vspace*{-5mm}

\vspace*{-5mm}
\begin{table}[H]
\centering
\vspace*{-2mm}
\caption{Proposed benchmark datasets$^*$ (test set 20\%). For the test sets, the total number of cases are listed first, followed by cases at the beginning of the test set missing some of their shorter prefixes (red/grey in Fig.~\ref{fig:traintest}), cases at the end of the test set missing their longer prefixes (grey/black in Fig.~\ref{fig:traintest}) and cases with all prefixes present in the test sets (grey in Fig.~\ref{fig:traintest}). The last column is the total number of cases in the dataset, but with the average of both kinds of incomplete test set cases rather than their sum.}\label{tab:proposed1}
\vspace*{-1mm}
\scalebox{0.999}{
\begin{tabular}{|l||c||c|c:c:c:c|c:c|}
\hline
\multirow{4}*{Dataset} &\multicolumn{8}{c|}{Number of cases}\\
\cline{2-9}
                     &Original           &Training  &\multicolumn{4}{c|}{Test}  &\multicolumn{2}{c|}{Total}    \\
&dataset                    &  set                 &\multicolumn{4}{c|}{set} &\multicolumn{2}{c|}{dataset}   \\
\cline{4-9}
                       &     &      &all      & missing& missing  & com-      &all  & full     \\
                       &                      &          & &short &long  &plete&          & case    \\
 & & & & prefixes& prefixes& && equivalent    \\                       
            
\hline
BPIC\_2012 & 12,183& 7.019& 2,468&570 &621&1,340&9,487&8,955  \\
BPIC\_2015 &5,641 &3,311 &1,116 &229 &221 & 666&4,427&4,202   \\
BPIC\_2017 &31,497 &21,404 &7,902 &2,040 &1,754 &4,108 &29,306 & 27,409  \\
BPIC\_2019 &251,465 & 84,233&86,905& 52,676& 46,423& 5,327& 171,138& 139,110 \\
\hdashline
BPIC\_2020:& & & & & &  & & \\
Dom. Decl.&6,087 &53& 522&407 &479 & 12 & 575& 508  \\
Intl. Decl.&1,440 &- & -&-& -&-  &- &  -\\
Payments&6,593 &4,494 &1,293 &135 & 171& 987& 5,787& 5,634  \\
Permits&7,063 &4,168 &2,679 & 1,309& 831& 667& 6,847&5,905   \\
Travel Costs& 2,097&1,265 & 533& 173&100 &260 & 1,798&1,662  \\

\hline
\multicolumn{9}{l}{\small *Outliers removed as per Subsection~\ref{subsec:chronoutliers}.} \\
\end{tabular}
}
\vspace*{-1mm}
\end{table}
\vspace*{-5mm}

\vspace*{-1mm}
\begin{table}[h]
\centering
\vspace*{-1mm}
\caption{Proposed benchmark datasets$^*$ (test set 20\%): Timing.}\label{tab:proposed2}
\scalebox{0.999}{
\begin{tabular}{|l||c|c|c|c||c|c|c|c|}
\hline
\multirow{4}*{Dataset} &\multicolumn{4}{c||}{Original datasets$^*$}&\multicolumn{4}{c|}{Proposed datasets}\\
\cline{2-9}
                       &Max.           &Data-   & Data-   & Data- & Max.   & Data-   &   Test & Test \\
                       &duration       &set     &set      &set    &dura-   & set     & set    & (= Data-)       \\
                       &cases          &nr.     &start    &end    &tion    & nr.     & start  & set        \\
                       &               &days    &         &       &cases   & days    &        &  end\\                       
            
\hline
BPIC\_2012  &137.2&152&2011-10&2012-02 &32.3&119.7&2012-01-05&2012-01-28  \\
BPIC\_2015 &1,512 &1,617&2010-10&2015-03 &302.8&1,314.2&2013-06-10&2014-05-10  \\
BPIC\_2017 &286.1 &397&2016-01& 2017-01&47.8&348.5&2016-10-10&2016-12-14  \\
BPIC\_2019 &380.0 &830&2018-01&2019-02 &143.3&253.6&2018-07-19&2018-09-11 \\
\hdashline
BPIC\_2020:& & & & & & & & \\
Dom. Decl.& 735.2 &773&2017-01&2019-02 &463.0&310.3&2017-09-05&2017-11-15  \\
Intl. Decl. &781.7 &1,010&2016-10&2019-07&-&-&-& -\\
Payments&1,238 &711&2017-01& 2018-12&28.9 &682.2&2018-09-14&2018-11-22 \\
Permits&502.4 &1,121&2016-10&2019-10 &258.8 &806.2&2018-09-07&2018-12-20 \\
Travel Costs&325.0 &749&2017-01&2019-01 & 114.3&634.0&2018-07-06&2018-10-06 \\

\hline
\multicolumn{9}{l}{\small *Outliers removed as per Subsection~\ref{subsec:chronoutliers}.} \\
\end{tabular}
}
\vspace*{-1mm}
\end{table}
\vspace*{-3mm}

\vspace*{-1mm}


\section{Conclusion and future work}\label{sec:conclusions}
In this paper, we demonstrated the impact of dataset preprocessing choices and pointed out the problem of overlapping training and test sets and at two sources of bias in test sets. Left unaddressed, these three obstacles risk slowing down progress in the field of predictive process monitoring. To overcome this challenge, we argued for the use of unbiased benchmark datasets with strict temporal splitting of training and test sets. Using the example of nine BPIC datasets and the remaining time prediction problem, we established preprocessing steps to construct such benchmark datasets. The scripts for our examples are publicly available. It is our hope that researchers will use these benchmarks rather than distil their own training and test sets from the original datasets and, by doing so, facilitate a more rapid advancement of predictive process monitoring. We also hope that other publicly available datasets will be constructed using our proposed transformation steps. The scientific marketplace will then decide which of these datasets will become predictive process monitoring's next MNIST, CIFAR-10 or ImageNet.


\end{document}